\documentclass[10pt,twocolumn,letterpaper]{article}

\usepackage{iccv}
\usepackage{times}
\usepackage{epsfig}
\usepackage{graphicx}
\usepackage{amsmath}
\usepackage{amssymb}

\usepackage{float}
\usepackage{booktabs}
\usepackage{caption}
\usepackage{subcaption}
\usepackage{multirow}
\usepackage{xcolor}
\usepackage{cite}
\usepackage[utf8]{inputenc}
\usepackage{url}
\usepackage[accsupp]{axessibility}  

\newcommand{\ba}{\mathbf{a}}

\newcommand{\bc}{\mathbf{c}}

\newcommand{\bo}{\mathbf{o}}
\newcommand{\bp}{\mathbf{p}}

\newcommand{\br}{\mathbf{r}}
\newcommand{\bs}{\mathbf{s}}

\newcommand{\bw}{\mathbf{w}}
\newcommand{\bx}{\mathbf{x}}

\newcommand{\nR}{\mathbb{R}}

\newcommand{\cL}{\mathcal{L}}

\newcommand{\cX}{\mathcal{X}}

\newcommand{\figref}[1]{Fig.~\ref{#1}}

\newcommand{\tabref}[1]{Table~\ref{#1}}

\makeatletter
\DeclareRobustCommand\onedot{\futurelet\@let@token\@onedot}
\def\@onedot{\ifx\@let@token.\else.\null\fi\xspace}
\def\eg{e.g\onedot} 
\def\ie{i.e\onedot}

\makeatother

\newcommand{\boldparagraph}[1]{\vspace{0.1cm}\noindent{\bf #1:}}

\definecolor{darkgreen}{rgb}{0,0.7,0}
\newcommand\crule[3][black]{\textcolor[HTML]{#1}{\rule{#2}{#3}}}

\graphicspath{{gfx/}}

\usepackage[pagebackref=true,breaklinks=true,letterpaper=true,colorlinks,bookmarks=false]{hyperref}

\iccvfinalcopy 

\def\iccvPaperID{3370} 

\ificcvfinal\fi

\begin{document}

\title{NEAT: Neural Attention Fields for End-to-End Autonomous Driving}

\author{Kashyap Chitta\thanks{indicates equal contribution} $^{1,2}$ \quad \quad Aditya Prakash\footnotemark[1] $^{1}$ \quad \quad Andreas Geiger$^{1,2}$\\
$^{1}$Max Planck Institute for Intelligent Systems, T\"ubingen \quad \quad $^{2}$University of T\"ubingen\\
{\tt\small \{firstname.lastname\}@tue.mpg.de}
}

\maketitle
\ificcvfinal\thispagestyle{empty}\fi

\begin{abstract}
Efficient reasoning about the semantic, spatial, and temporal structure of a scene is a crucial prerequisite for autonomous driving. We present NEural ATtention fields (NEAT), a novel representation that enables such reasoning for end-to-end imitation learning models. NEAT is a continuous function which maps locations in Bird's Eye View (BEV) scene coordinates to waypoints and semantics, using intermediate attention maps to iteratively compress high-dimensional 2D image features into a compact representation. This allows our model to selectively attend to relevant regions in the input while ignoring information irrelevant to the driving task, effectively associating the images with the BEV representation. In a new evaluation setting involving adverse environmental conditions and challenging scenarios, NEAT outperforms several strong baselines and achieves driving scores on par with the privileged CARLA expert used to generate its training data. Furthermore, visualizing the attention maps for models with NEAT intermediate representations provides improved interpretability.
\end{abstract}
\vspace{-0.4cm}
\section{Introduction}
\label{sec:intro}

\begin{figure}[t!]
\centering
\includegraphics[width=\columnwidth]{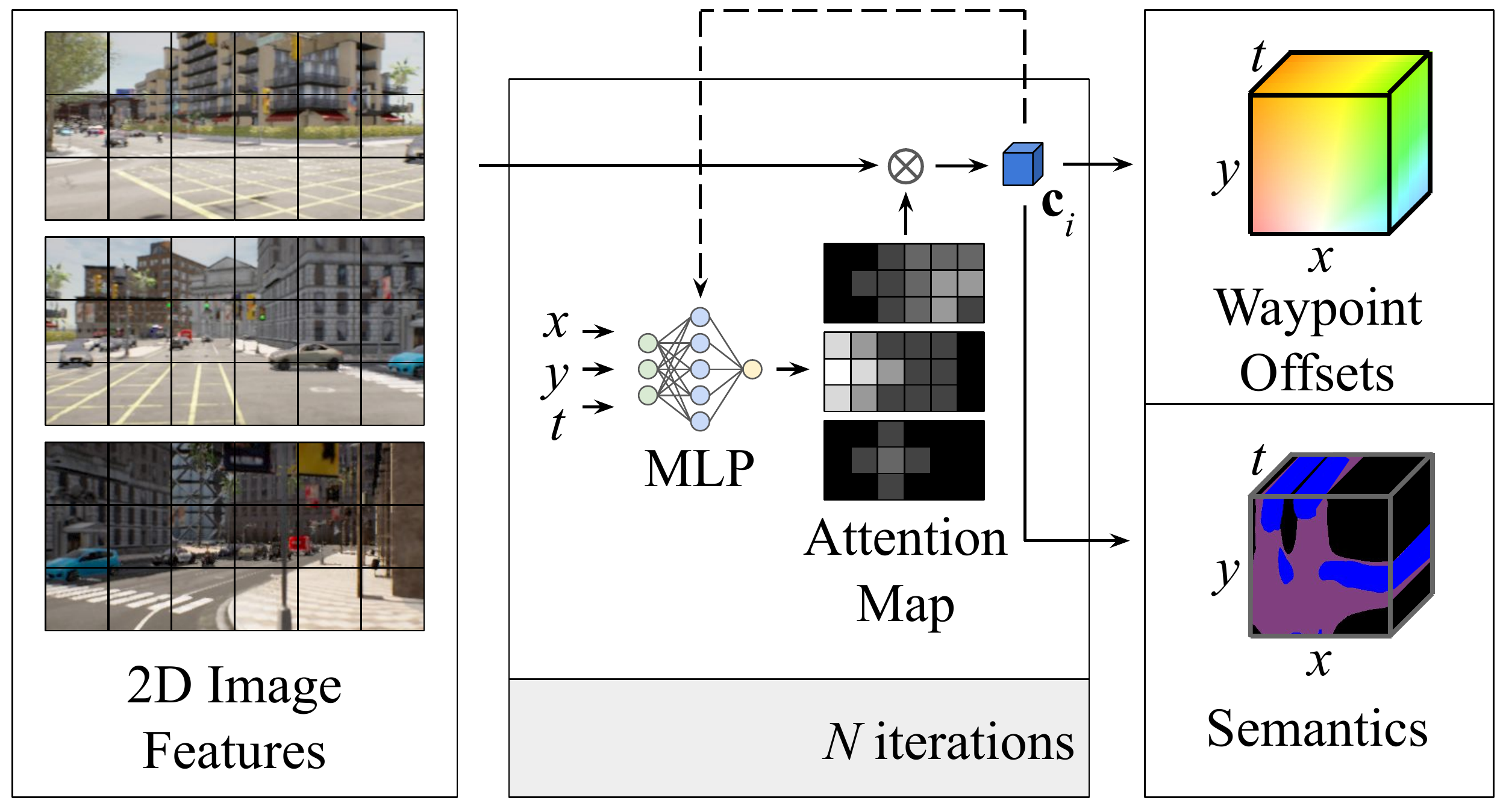}
\caption{\textbf{Neural Attention Fields.} We use an MLP to iteratively compress the high-dimensional input into a compact low-dimensional representation $\bc_i$ based on the BEV query location $(x,y,t)$. Our model outputs waypoint offsets and auxiliary semantics from $\bc_i$ continuously and with a low memory footprint. Training for both tasks jointly leads to improved driving performance on CARLA.}
\label{fig:teaser}
\vspace{-0.4cm}
\end{figure}

Navigating large dynamic scenes for autonomous driving requires a meaningful representation of both the spatial and temporal aspects of the scene. Imitation Learning (IL) by behavior cloning has emerged as a promising approach for this task~\cite{Pomerleau1988NIPS,Bojarski2016ARXIV,Codevilla2019ICCV,Zeng2019CVPR,Chen2019CORL}. Given a dataset of expert trajectories, a behavior cloning agent is trained through supervised learning, where the goal is to predict the actions of the expert given some sensory input regarding the scene~\cite{Osa2018}. To account for the complex spatial and temporal scene structure encountered in autonomous driving, the training objectives used in IL-based driving agents have evolved by incorporating auxiliary tasks. Pioneering methods, such as CILRS~\cite{Codevilla2019ICCV}, use a simple self-supervised auxiliary training objective of predicting the ego-vehicle velocity. Since then, more complex training signals aiming to reconstruct the scene have become common, \eg image auto-encoding~\cite{Ohn-Bar2020CVPR}, 2D semantic segmentation~\cite{Huang2020SENSORS}, Bird's Eye View (BEV) semantic segmentation~\cite{Loukkal2020ARXIV}, 2D semantic prediction~\cite{Hu2020ECCV}, and BEV semantic prediction~\cite{Sadat2020ECCV}. Performing an auxiliary task such as BEV semantic prediction, which requires the model to output the BEV semantic segmentation of the scene at both the observed and future time-steps, incorporates spatiotemporal structure into the intermediate representations learned by the agent. This has been shown to lead to more interpretable and robust models~\cite{Sadat2020ECCV}. However, so far, this has only been possible with expensive LiDAR and HD map-based network inputs which can be easily projected into the BEV coordinate frame. 

The key challenge impeding BEV semantic prediction from camera inputs is one of association: given a BEV spatiotemporal query location $(x,y,t)$ in the scene (\eg 2 meters in front of the vehicle, 5 meters to the right, and 2 seconds into the future), it is difficult to identify which image pixels to associate to this location, as this requires reasoning about 3D geometry, scene motion, ego-motion, and intention, as well as interactions between scene elements. In this paper, we propose \textbf{NEural ATtention fields (NEAT)}, a flexible and efficient feature representation designed to address this challenge. Inspired by implicit shape representations~\cite{Mescheder2019CVPR,Park2019CVPR}, NEAT represents large dynamic scenes with a fixed memory footprint using a multi-layer perceptron (MLP) query function. The core idea is to learn a function from any query location $(x,y,t)$ to an attention map for features obtained by encoding the input images. NEAT compresses the high-dimensional image features into a compact low-dimensional representation relevant to the query location $(x,y,t)$, and provides interpretable attention maps as part of this process, without attention supervision~\cite{Zhang2018ECCVb}. As shown in \figref{fig:teaser}, the output of this learned MLP can be used for dense prediction in space and time. Our end-to-end approach predicts waypoint offsets to solve the main trajectory planning task (described in detail in Section \ref{sec:method}), and uses BEV semantic prediction as an auxiliary task.

Using NEAT intermediate representations, we train several autonomous driving models for the CARLA driving simulator~\cite{Dosovitskiy2017CORL}. We consider a more challenging evaluation setting than existing work based on the new CARLA Leaderboard~\cite{Leaderboard} with CARLA version 0.9.10, involving the presence of multiple evaluation towns, new environmental conditions, and challenging pre-crash traffic scenarios. We outperform several strong baselines and match the privileged expert's performance on our internal evaluation routes. On the secret routes of the CARLA Leaderboard, NEAT obtains competitive driving scores while incurring significantly fewer infractions than existing methods.

\boldparagraph{Contributions} (1) We propose an architecture combining our novel NEAT feature representation with an implicit decoder~\cite{Mescheder2019CVPR} for joint trajectory planning and BEV semantic prediction in autonomous vehicles. (2) We design a challenging new evaluation setting in CARLA consisting of 6 towns and 42 environmental conditions and conduct a detailed empirical analysis to demonstrate the driving performance of NEAT. (3) We visualize attention maps and semantic scene interpolations from our interpretable model, yielding insights into the learned driving behavior. Our code is available at \url{https://github.com/autonomousvision/neat}.
\section{Related Work}
\label{sec:related}

\boldparagraph{Implicit Scene Representations} The geometric deep learning community has pioneered the idea of using neural implicit representations of scene geometry. These methods represent surfaces as the boundary of a neural classifier~\cite{Mescheder2019CVPR,Chen2019CVPR,Saito2019ICCV,Liu2019NIPSb,Chibane2020CVPR} or zero-level set of a signed distance field regression function~\cite{Michalkiewicz2019ICCV,Park2019CVPR,Xu2019NIPS,Liu2020CVPR,Sitzmann2020NIPSa,Sitzmann2020NIPSb}. They have been applied for representing object texture \cite{Oechsle2019ICCV,Sitzmann2019NIPS,Niemeyer2020CVPR}, dynamics~\cite{Niemeyer2019ICCV} and lighting properties~\cite{Mildenhall2020ECCV,Oechsle2020THREEDV,Schwarz2020NIPS}. Recently, there has been progress in applying these representations to compose objects from primitives~\cite{Genova2019ICCV,Deng2020CVPR,Chen2020CVPR,Deng2020ECCV}, and to represent larger scenes, both static~\cite{Peng2020ECCV,Jiang2020CVPR,Chabra2020ECCV} and dynamic~\cite{Li2020ARXIV,Li2021ARXIV,Du2020ARXIV,Xian2020ARXIV}. These methods obtain high-resolution scene representations while remaining compact, due to the constant memory footprint of the neural function approximator. While NEAT is motivated by the same property, we use the compactness of neural approximators to learn better intermediate features for the downstream driving task.

\boldparagraph{End-to-End Autonomous Driving} Learning-based autonomous driving is an active research area~\cite{Tampuu2020ARXIV,Janai2020}. IL for driving has advanced significantly~\cite{Pomerleau1988NIPS,Bojarski2016ARXIV,Codevilla2018ICRA,Muller2018CORL,Zeng2019CVPR,Wei2020ARXIV} and is currently employed in several state-of-the-art approaches, some of which predict waypoints~\cite{Chen2019CORL,Filos2020ICML,Casas2021ARXIV}, whereas others directly predict vehicular control~\cite{Codevilla2019ICCV,Prakash2020CVPR,Ohn-Bar2020CVPR,Huang2020SENSORS,Xiao2020TITS,Behl2020IROS,Buhler2020IROS,Zhao2020CORL}. While other learning-based driving methods such as affordances~\cite{Sauer2018CORL,Xiao2020CORL} and Reinforcement Learning~\cite{Toromanoff2020CVPR,Wang2020ARXIV,Chen2021ARXIV} could also benefit from a NEAT-based encoder, in this work, we apply NEAT to improve IL-based autonomous driving. 

\boldparagraph{BEV Semantics for Driving} A top-down view of a street scene is powerful for learning the driving task since it contains information regarding the 3D scene layout, objects do not occlude each other, and it represents an orthographic projection of the physical 3D space which is better correlated with vehicle kinematics than the projective 2D image domain. LBC~\cite{Chen2019CORL} exploits this representation in a teacher-student approach. A teacher that learns to drive given BEV semantic inputs is used to supervise a student aiming to perform the same task from images only. By doing so, LBC achieves state-of-the-art performance on the previous CARLA version 0.9.6, showcasing the benefits of the BEV representation. NEAT differs from LBC by directly learning in BEV space, unlike the LBC student model which learns a classical image-to-trajectory mapping.

Other works deal with BEV scenes, \eg, obtaining BEV projections~\cite{Zhu2018THREEDV,Abbas2019ICCVWORK} or BEV semantic predictions~\cite{Mani2020WACV,Roddick2020CVPR,Pan2020RAL,Hendy2020ARXIV,Hu2021ARXIV} from images, but do not use these predictions for driving. More recently, LSS~\cite{Philion2020ECCV} and OGMs~\cite{Loukkal2020ARXIV} demonstrated joint BEV semantic reconstruction and driving from camera inputs. Both methods involve explicit projection based on camera intrinsics, unlike our learned attention-based feature association. They only predict semantics for static scenes, while our model includes a time component, performing prediction up to a fixed horizon. Moreover, unlike us, they only evaluate using offline metrics which are known to not necessarily correlate well with actual downstream driving performance~\cite{Codevilla2018ECCV}. Another related work is P3~\cite{Sadat2020ECCV} which jointly performs BEV semantic prediction and driving. In comparison to P3 which uses expensive LiDAR and HD map inputs, we focus on image modalities.
\section{Method}
\label{sec:method}

A common approach to learning the driving task from expert demonstrations is end-to-end trajectory planning, which uses \textit{waypoints} $\bw_t$ as outputs. A waypoint is defined as the position of the vehicle in the expert demonstration at time-step $t$, in a BEV projection of the vehicle's local coordinate system. The coordinate axes are fixed such that the vehicle is located at $(x,y)=(0,0)$ at the current time-step $t=T$, and the front of the vehicle is aligned along the positive y-axis. Waypoints from a sequence of future time-steps $t=T+1,...,T+Z$ form a trajectory that can be used to control the vehicle, where $Z$ is a fixed prediction horizon.

As our agent drives through the scene, we collect sensor data into a fixed-length buffer of $T$ time-steps, $\cX=\{\bx_{s,t}\}_{s=1:S, t=1:T}$ where each $\bx_{s,t}$ comes from one of $S$ sensors. The final frame in the buffer is always the current time-step ($t=T$). In practice, the $S$ sensors are RGB cameras, the standard input modality in existing work on CARLA~\cite{Chen2019CORL}. By default, we use $S=3$ cameras, one oriented forward and the others 60 degrees to the left and right. After cropping these camera images to remove radial distortion, these $S=3$ images together provide a full 180$^\circ$ view of the scene in front of the vehicle. While NEAT can be applied with different buffer sizes, we focus in our experiments on the setting where the input is a single frame ($T=1$), as several studies indicate that using historical observations can be detrimental to the driving task~\cite{Wang2019IROS,Wen2020NIPS}.

In addition to waypoints, we use BEV semantic prediction as an auxiliary task to improve driving performance. Unlike waypoints which are small in number (\eg $Z=4$) and can be predicted discretely, BEV semantic prediction is a dense prediction task, aiming to predict semantic labels at any spatiotemporal query location $(x,y,t)$ bounded to some spatial range and the time interval $1\le t\le T+Z$. Predicting both observed ($1\le t<T$) and future ($T<t\le T+Z$) semantics provides a holistic understanding of the scene dynamics. Dynamics prediction from a single input frame is possible since the orientation and position of vehicles encodes information regarding their motion~\cite{Walker2015ICCV}.

\begin{figure}[t!]
\centering
\includegraphics[width=\columnwidth]{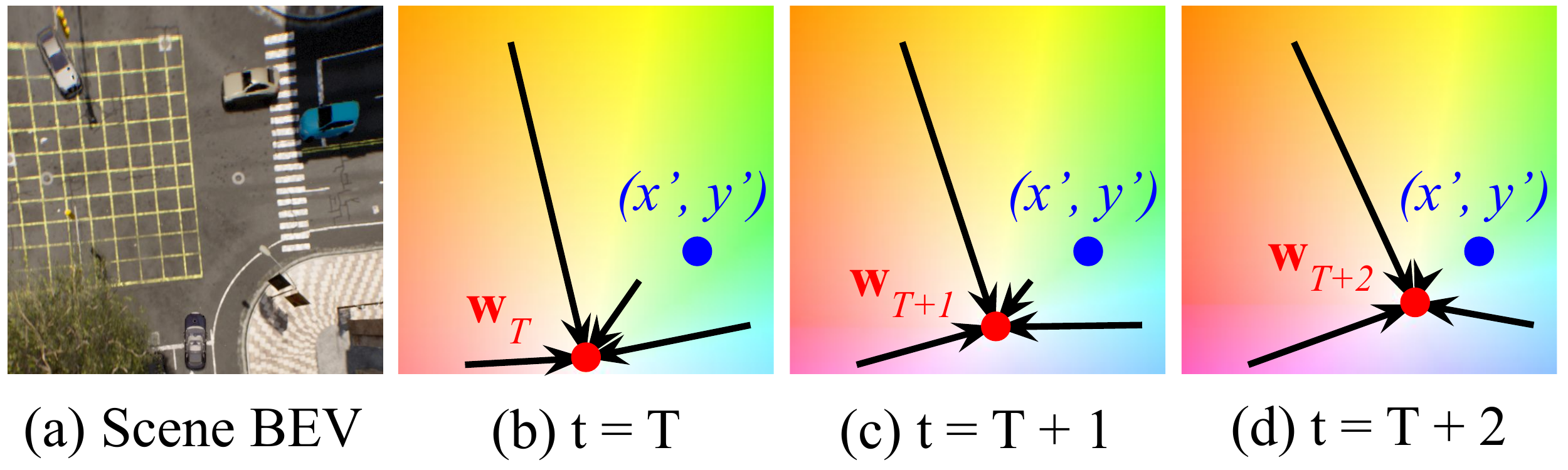}
\caption{\textbf{Dense offset prediction.} We visualize the target location $(x',y')$ ({\color{blue}blue dot}), waypoint $\bw_t$ ({\color{red}red dot}) and waypoint offsets $\bo$ (arrows) for a scene at three time instants. The offsets $\bo$ represent the 2D vector from any query location $(x,y)$ to the waypoint $\bw_t$ at time $t$ and thus implicitly represent the waypoint. The arrows illustrate $\bo$ for four different query locations $(x,y)$. We also show a color coding based visualization of the dense flow field learned by our model, representing $\bo$ from any $(x,y)$ location in the scene.}
\label{fig:bev}
\vspace{-0.3cm}
\end{figure}

The coordinate system used for BEV semantic prediction is the same as the one used for waypoints. Thus, if we frame waypoint prediction as a dense prediction task, it can be solved simultaneously with BEV semantic prediction using the proposed NEAT as a shared representation. Therefore, we propose a \textbf{dense offset prediction} task to locate waypoints as visualized in \figref{fig:bev} using a standard optical flow color wheel~\cite{Baker2011IJCV}. The goal is to learn the field of 2-dimensional offset vectors $\bo$ from query locations $(x,y,t)$ to the {waypoint} $\bw_t$ (\eg $\bo=(0,0)$ when $(x,y)=\bw_T$ and  $t=T$). In certain situations, future waypoints along different trajectories are plausible (\eg taking a left or right turn at an intersection), thus it is important to adapt $\bo$ based on the driver intention. We do this by using provided \textit{target locations} $(x',y')$ as inputs. Target locations are GPS coordinates provided by a navigational system along the route to be followed. They are transformed to the same coordinate system as the waypoints before being used as inputs. These target locations are sparse and can be hundreds of meters apart. In \figref{fig:bev}, the target location to the right of the intersection helps the model decide to turn right rather than proceeding straight. We choose target locations as the method for specifying driver intention as they are the default intention signal in the CARLA simulator since version 0.9.9. In summary, the goal of dense offset prediction is to output $\bo$ for any 5-dimensional query point $\bp=(x,y,t,x',y')$.

\subsection{Architecture}

\begin{figure*}[t!]
\centering
\includegraphics[width=\textwidth]{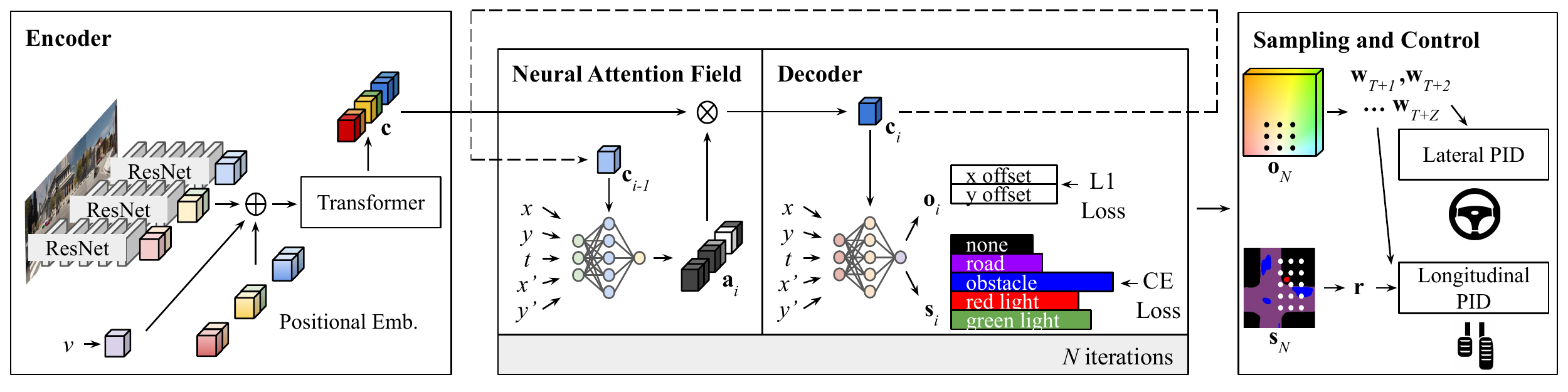}
\caption{\textbf{Model Overview.} In the encoder, image patch features, velocity features, and a learned positional embedding are summed and fed into a transformer. We illustrate this with 2 features per image, though our model uses 64 in practice. NEAT recurrently updates an attention map $\ba_i$ for the encoded features $\bc$ for $N$ iterations. The inputs to NEAT are a query point $\bp=(x,y,t,x',y')$ and feature $\bc_i$. For the initial iteration, $\bc_0$ is set to the mean of $\bc$. The dotted arrow shows the recursion of features between subsequent iterations. In each iteration, the decoder predicts the waypoint offset $\bo_i$ and the semantic class $\bs_i$ for any given query $\bp$, which are supervised using loss functions. At test time, we sample predictions from grids on $\bo_N$ and $\bs_N$ to obtain a waypoint for each time-step $\bw_t$ and red light indicator $\br$, which are used by PID controllers for driving.}
\label{fig:arch}
\vspace{-0.3cm}
\end{figure*}

As illustrated in \figref{fig:arch}, our architecture consists of three neural networks that are jointly trained for the BEV semantic prediction and dense offset prediction tasks: an encoder $e_\theta$, neural attention field $a_\phi$, and decoder $d_\psi$. In the following, we go over each of the three components in detail.

\boldparagraph{Encoder}
Our encoder $e_\theta$ takes as inputs the sensor data buffer $\cX$ and a scalar $v$, which is the vehicle velocity at the current time-step $T$. Formally, it is denoted as
\begin{equation}
    e_\theta : \nR^{S\times T\times W\times H\times 3} \times \nR \rightarrow \nR^{(S*T*P) \times C}
\end{equation}
where $\theta$ denotes the encoder parameters. Each image $\bx_{s,t}\in \nR^{W\times H\times 3}$ is processed by a ResNet \cite{He2016CVPR} to provide a grid of features from the penultimate layer of size $\nR^{P\times C}$, where $P$ is the number of spatial features per image and $C$ is the feature dimensionality. For the $256\times 256$ pixel input resolution we consider, we obtain $P=64$ patches from the default ResNet architecture. These features are further processed by a transformer~\cite{Vaswani2017NIPS}. The goal of the transformer is to integrate features globally, adding contextual cues to each patch with its self-attention mechanism. This enables interactions between features across different images and over a large spatial range. Note that the transformer can be removed from our encoder without changing the output dimensionality, but we include it since it provides an improvement as per our ablation study. Before being input to the transformer, each patch feature is combined (through addition) with (1) a velocity feature obtained by linearly projecting $v$ to $\nR^C$, and broadcasting to all patches of all sensors at all time-steps, as well as (2) a learned positional embedding, which is a trainable parameter of size $(S*T*P) \times C$. The transformer outputs patch features $\bc \in \nR^{(S*T*P) \times C}$.

\boldparagraph{Neural Attention Field}
While the transformer aggregates features globally, it is not informed by the query and target location. Therefore, we introduce NEAT (\figref{fig:teaser}), which identifies the patch features from the encoder relevant for making predictions regarding any query point in the scene $\bp=(x,y,t,x',y')$. It introduces a bottleneck in the network and improves interpretability (\figref{fig:att}). Its operation can be formally described as
\begin{equation}
    a_\phi : \nR^5 \times \nR^C \rightarrow \nR^{S*T*P}
\end{equation}
Note that the target location $(x',y')$ input to NEAT is omitted in \figref{fig:teaser} for clarity. While NEAT could in principle directly take as inputs $\bp$ and $\bc$, this would be inefficient due to the high dimensionality of $\bc \in \nR^{(S*T*P) \times C}$. We instead use a simple iterative attention process with $N$ iterations. At iteration $i$, the output $\ba_i \in \nR^{S*T*P}$ of NEAT is used to obtain a feature $\bc_i \in \nR^C$ specific to the query point $\bp$ through a softmax-scaled dot product between $\ba_i$ and $\bc$:
\begin{equation}
    \bc_i = \text{softmax}(\ba_i)^\top \cdot \bc
    \label{eqn:softmax}
\end{equation}
The feature $\bc_i$ is used as the input of $a_\phi$ along with $\bp$ at the next attention iteration, implementing a recurrent attention loop (see \figref{fig:arch}). Note that the dimensionality of $\bc_i$ is significantly smaller than that of the transformer output $\bc$, as $\bc_i$ aggregates information (via Eq. \eqref{eqn:softmax}) across sensors $S$, time-steps $T$ and patches $P$. For the initial iteration, $\bc_0$ is set to the mean of $\bc$ (equivalent to assuming a uniform initial attention). We implement $a_\phi$ as a fully-connected MLP with 5 ResNet blocks of 128 hidden units each, conditioned on $\bc_i$ using conditional batch normalization \cite{Vries2017NIPS,Dumoulin2017ICLR} (details in supplementary). We share the weights of $a_\phi$ across all iterations which works well in practice.

\boldparagraph{Decoder}
The final network in our model is the decoder:
\begin{equation}
    d_\psi: \nR^5 \times \nR^C \rightarrow \nR^K \times \nR^2
\end{equation}
It is an MLP with a similar structure to $a_\phi$, but differing in terms of its output layers. Given $\bp$ and $\bc_i$, the decoder predicts the semantic class $\bs_i \in \nR^K$ (where $K$ is the number of classes) and waypoint offset $\bo_i \in \nR^2$ at each of the $N$ attention iterations. While the outputs decoded at intermediate iterations ($i<N$) are not used at test time, we supervise these predictions during training to ease optimization.

\subsection{Training}

\boldparagraph{Sampling}
An important consideration is the choice of query samples $\bp$ during training, and how to acquire ground truth labels for these points. Among the 5 dimensions of $\bp$, $x'$ and $y'$ are fixed for any $\cX$, but $x$, $y$, and $t$ can all be varied to access different positions in the scene. Note that in the CARLA simulator, the ground truth waypoint is only available at discrete time-steps, and the ground truth semantic class only at discrete $(x,y,t)$ locations. However, this is not an issue for NEAT as we can supervise our model using arbitrarily sparse observations in the space-time volume. We consider $K=5$ semantic classes by default: none, road, obstacle (person or vehicle), red light, and green light. The location and state of the traffic light affecting the ego-vehicle are provided by CARLA. We use this to set the semantic label for points within a fixed radius of the traffic light pole to the red light or green light class, similar to~\cite{Chen2019CORL}. In our work, we focus on the simulated setting where this information is readily available, though BEV semantic annotations of objects (obstacles, red lights, and green lights) can also be obtained for real scenes using projection. The only remaining labels required by NEAT (for the road class) can be obtained by fitting the ground plane to LiDAR sweeps in a real dataset or more directly from localized HD maps.

We acquire these BEV semantic annotations from CARLA up to a fixed prediction horizon $Z$ after the current time-step $T$ and register them to the coordinate frame of the ego-vehicle at $t=T$. $Z$ is a hyper-parameter that can be used to modulate the difficulty of the prediction task. From the aligned BEV semantic images, we only consider points approximately in the field-of-view of our camera sensors. We use a range of 50 meters in front of the vehicle and 25 meters to either side (detailed sensor configurations are provided in the supplementary).

Since semantic class labels are typically heavily imbalanced, simply using all observations for training (or sampling a random subset) would lead to several semantic classes being under-represented in the training distribution. We use a class-balancing sampling heuristic during training to counter this. To sample $M$ points for $K$ semantic classes, we first group all the points from all the $T+Z$ time-steps in the sequence into bins based on their semantic label. We then attempt to randomly draw $\frac{M}{K}$ points from each bin, starting with the class having the least number of available points. If we are unable to sample enough points from any class, this difference is instead sampled from the next bin, always prioritizing classes with fewer total points.

We obtain the offset ground truth for each of these $M$ sampled points by collecting the ground truth waypoints for the $T+Z$ time-steps around each frame. The offset label for each of the $M$ sampled points is calculated as its difference from the ground truth waypoint at the corresponding time-step $\bw_t$. Being a regression task, we find that offset prediction does not benefit as much from a specialized sampling strategy. Therefore, we use the same $M$ points for supervising the offsets even though they are sampled based on semantic class imbalance, improving training efficiency. 

\boldparagraph{Loss} For each of the $M$ sampled points, the decoder makes predictions $\bs_i$ and $\bo_i$ at each of the $N$ attention iterations. The encoder, NEAT and decoder are trained jointly with a loss function applied to these $MN$ predictions:
\begin{equation}
\label{eqn:loss}
    \cL = \frac{1}{MN} \sum_{i}^{N} \gamma_i \sum_{j}^{M} {||\bo^*_j - \bo_{i,j}||}_1 +  \lambda \cL_{CE}(\bs^*_j, \bs_{i,j})
\end{equation}
where ${||.||}_1$ is the $L_1$ distance between the true offset $\bo^*$ and predicted offset $\bo_i$, $\cL_{CE}$ is the cross-entropy between the true semantic class $\bs^*$ and predicted class $\bs_i$, $\lambda$ is a weight between the semantic and offset terms, and $\gamma_i$ is used to down-weight predictions made at earlier iterations ($i<N$). These intermediate losses improve performance, as we show in our experiments.

\subsection{Controller}

To drive the vehicle at test time, we generate a red light indicator and waypoints from our trained model; and convert them into steer, throttle, and brake values. For the red light indicator, we uniformly sample a sparse grid of $U\times V$ points in $(x,y)$ at the current time-step $t=T$, in the area 50 meters to the front and 25 meters to the right side of the vehicle. We append the target location $(x',y')$ to these grid samples to obtain 5-dimensional queries that can be used as NEAT inputs. From the semantic prediction obtained for these points at the final attention iteration $\bs_N$, we set the red light indicator $\br$ as 0 if none of the points belongs to the red light class, and 1 otherwise. In our ablation study, we find this indicator to be important for performance.

To generate waypoints, we sample a uniformly spaced grid of $G\times G$ points in a square region of side $A$ meters centered at the ego-vehicle at each of the future time-steps $t=T+1,\dots,T+Z$. Note that predicting waypoints with a single query point ($G=1$) is possible, but we use a grid for robustness. After encoding the sensor data and performing $N$ attention iterations, we obtain $\bo_N$ for each of the $G^2$ query points at each of the $Z$ future time-steps. We offset the $(x,y)$ location coordinates of each query point towards the waypoint by adding $\bo_N$, effectively obtaining the waypoint prediction for that sample, \ie $(\bp[0],\bp[1]) \texttt{\,+=\,} \bo_N$. After this offset operation, we average all $G^2$ waypoint predictions at each future time instant, yielding the final waypoint predictions $\bw_t$. To obtain the throttle and brake values, we compute the vectors between waypoints of consecutive time-steps and input the magnitude of these vectors to a longitudinal PID controller along with the red light indicator. The relative orientation of the waypoints is input to a lateral PID controller for turns. Please refer to the supplementary material for further details on both controllers.

\section{Experiments}
\label{sec:results}

In this section, we describe our experimental setting, showcase the \textbf{driving performance} of NEAT in comparison to several baselines, present an \textbf{ablation study} to highlight the importance of different components of our architecture, and show the interpretability of our approach through \textbf{visualizations} obtained from our trained model.

\boldparagraph{Task}
We consider the task of navigation along pre-defined routes in CARLA version 0.9.10~\cite{Dosovitskiy2017CORL}. A route is defined by a sequence of sparse GPS locations (target locations). The agent needs to complete the route while coping with background dynamic agents (pedestrians, cyclists, vehicles) and following traffic rules. We tackle a new challenge in CARLA 0.9.10: each of our routes may contain several pre-defined dangerous \textbf{scenarios} (\eg unprotected turns, other vehicles running red lights, pedestrians emerging from occluded regions to cross the road).

\boldparagraph{Routes}
For training data generation, we store data using an expert agent along routes from the 8 publicly available towns in CARLA, randomly spawning scenarios at several locations along each route. We evaluate NEAT on the official CARLA Leaderboard~\cite{Leaderboard}, which consists of 100 secret routes with unknown environmental conditions. We additionally conduct an internal evaluation consisting of 42 routes from 6 different CARLA towns (Town01-Town06). Each route has a unique environmental condition combining one of 7 weather conditions (Clear, Cloudy, Wet, MidRain, WetCloudy, HardRain, SoftRain) with one of 6 daylight conditions (Night, Twilight, Dawn, Morning, Noon, Sunset). Additional details regarding our training and evaluation routes are provided in the supplementary. Note that in this new evaluation setting, the multi-lane road layouts, distant traffic lights, high density of background agents, diverse daylight conditions, and new metrics which strongly penalize infractions (described below) make navigation more challenging, leading to reduced scores compared to previous CARLA benchmarks~\cite{Dosovitskiy2017CORL,Codevilla2019ICCV}.

\boldparagraph{Metrics}
We report the official metrics of the CARLA Leaderboard, \textbf{Route Completion (RC)}, \textbf{Infraction Score (IS)}\footnote{The Leaderboard refers to this as infraction penalty. We use the terminology `score' since it is a multiplier for which higher values are better.} and \textbf{Driving Score (DS)}. For a given route, RC is the percentage of the route distance completed by the agent before it deviates from the route or gets blocked. IS is a cumulative multiplicative penalty for every collision, lane infraction, red light violation, and stop sign violation. Please refer to the supplementary material for additional details regarding the penalty applied for each kind of infraction. Finally, DS is computed as the RC weighted by the IS for that route. After calculating all metrics per route, we report the mean performance over all 42 routes. We perform our internal evaluation three times for each model and report the mean and standard deviation for all metrics.

\begin{table*}[t]
    \begin{tabular}{c c}
        \begin{subtable}[h]{0.55\textwidth}
            \small
            \setlength{\tabcolsep}{5pt}
            \centering
            \begin{tabular}{c | c | c c c}
                \textbf{Method} & \textbf{Aux. Sup.} & \textbf{RC} $\uparrow$& \textbf{IS} $\uparrow$& \textbf{DS} $\uparrow$\\
                \hline
                CILRS~\cite{Codevilla2019ICCV} & Velocity & 35.46$\pm$0.41 & 0.66$\pm$0.02 & 22.97$\pm$0.90\\
                LBC~\cite{Chen2019CORL} & BEV Sem & 61.35$\pm$2.26 & 0.57$\pm$0.02 & 29.07$\pm$0.67 \\
                AIM~\cite{Prakash2021CVPR} & None & 70.04$\pm$2.31 & 0.73$\pm$0.03 & 51.25$\pm$0.17 \\
                \hline
                \multirow{3}{*}{AIM-MT} & 2D Sem & 80.21$\pm$3.55 & 0.74$\pm$0.02 & 57.95$\pm$2.76 \\
                 & BEV Sem & 77.93$\pm$3.06 & 0.78$\pm$0.01 & 60.62$\pm$2.33 \\
                 & Depth+2D Sem & \textbf{80.81$\pm$2.47} & 0.80$\pm$0.01 & 64.86$\pm$2.52 \\
                 \hline
                AIM-VA & 2D Sem & 75.40$\pm$1.53 & 0.79$\pm$0.02 & 60.94$\pm$0.79 \\
                \hline
                NEAT & BEV Sem & 79.17$\pm$3.25 & \textbf{0.82$\pm$0.01} & \textbf{65.10$\pm$1.75} \\
                \hline
                Expert & N/A & 86.05$\pm$2.58 & 0.76$\pm$0.01 & 62.69$\pm$2.36 \\
            \end{tabular}
            \caption{\textbf{CARLA 42 Routes.} We show the mean and standard deviation over 3 evaluations for each model. NEAT obtains the best driving score, on par with (and sometimes even outperforming) the expert used for data collection.}
            \label{tab:baselines}
            \vspace{0.0cm}
        \end{subtable}
        &
        \begin{subtable}[h]{0.40\textwidth}
            \small
            \setlength{\tabcolsep}{2.4pt}
        	\centering
        	\begin{tabular}{c | c | c | c c c}
                \textbf{\#} & \textbf{Method} & \textbf{Aux. Sup.} & \textbf{RC} $\uparrow$ & \textbf{IS} $\uparrow$ & \textbf{DS} $\uparrow$ \\
        		\hline
        		1 & WOR~\cite{Chen2021ARXIV} & 2D Sem & 57.65 & 0.56 & 31.37 \\
        		2 & MaRLn~\cite{Toromanoff2020CVPR} & 2D Sem+Aff & 46.97 & 0.52 & 24.98 \\
        		3 & NEAT (Ours) & BEV Sem & 41.71 & 0.65 & 21.83 \\
        		4 & AIM-MT & Depth+2D Sem & 67.02 & 0.39 & 19.38 \\
        		5 & TransFuser~\cite{Prakash2021CVPR} & None & 51.82 & 0.42 & 16.93 \\
        		6 & LBC~\cite{Chen2019CORL} & BEV Sem & 17.54 & 0.73 & 8.94 \\
        		7 & CILRS~\cite{Codevilla2019ICCV}& Velocity & 14.40 & 0.55 & 5.37 \\
        	\end{tabular}
        	\caption{\textbf{CARLA Leaderboard.} Among all publicly visible entries (accessed in July 2021), NEAT obtains the third-best DS. Of the top 3 methods, NEAT has the highest IS, indicating safer driving on unseen routes.}
        	\label{tab:challenge}
        	\vspace{0.0cm}
        \end{subtable}
    \end{tabular}
    \vspace{-0.2cm}
    \caption{\textbf{Quantitative Evaluation on CARLA.} We report the RC, IS and DS over our 42 internal evaluation routes (\tabref{tab:baselines}) and 100 secret routes on the evaluation server~\cite{Leaderboard} (\tabref{tab:challenge}). We abbreviate semantics with ``Sem'' and affordances with ``Aff''.}
    \label{tab:results}
    \vspace{-0.3cm}
\end{table*}

\boldparagraph{Baselines}
We compare our approach against several recent methods. \textbf{CILRS}~\cite{Codevilla2019ICCV} learns to directly predict vehicle controls (as opposed to waypoints) from visual features while being conditioned on a discrete navigational command (follow lane, change lane left/right, turn left/right). It is a widely used baseline for the old CARLA version 0.8.4, which we adapted to the latest CARLA version. \textbf{LBC}~\cite{Chen2019CORL} is a knowledge distillation approach where a teacher model with access to ground truth BEV semantic segmentation maps is first trained using expert supervision to predict waypoints, followed by an image-based student model which is trained using supervision from the teacher. It is the state-of-the-art approach on CARLA version 0.9.6. We train LBC on our dataset using the latest codebase provided by the authors for CARLA version 0.9.10. \textbf{AIM}~\cite{Prakash2021CVPR} is an improved version of CILRS, where a GRU decoder regresses waypoints. To assess the effects of different forms of auxiliary supervision, we create 3 multi-task variants of AIM (\textbf{AIM-MT}). Each variant adds a different auxiliary task during training: (1) 2D semantic segmentation using a deconvolutional decoder, (2) BEV semantic segmentation using a spatial broadcast decoder~\cite{Watters2019ICLRWORK}, and (3) both 2D depth estimation and 2D semantic segmentation as in~\cite{Li2018ARXIV}. We also replace the CILRS backbone of Visual Abstractions~\cite{Behl2020IROS} with AIM, to obtain \textbf{AIM-VA}. This approach generates 2D segmentation maps from its inputs which are then fed into the AIM model for driving. Finally, we report results for the privileged \textbf{Expert} used for generating our training data.

\boldparagraph{Implementation} By default, NEAT's transformer uses $L=2$ layers with 4 parallel attention heads. Unless otherwise specified, we use $T=1$, $Z=4$, $P=64$, $C=512$, $N=2$, $M=64$, $U=16$, $V=32$, $G=3$ and $A=2.5$. We use a weight of $\lambda=0.1$ on the $L_1$ loss, set $\gamma_{i}=0.1$ for the intermediate iterations ($i<N$), and set $\gamma_N=1$. For a fair comparison, we choose the best performing encoders for each model among ResNet-18, ResNet-34, and ResNet-50 (NEAT uses ResNet-34). Moreover, we chose the best out of two different camera configurations ($S=1$ and $S=3$) for each model, using a late fusion strategy for combining sensors in the baselines when we set $S=3$. Additional details are provided in the supplementary.

\subsection{Driving Performance}

Our results are presented in \tabref{tab:results}. \tabref{tab:baselines} focuses on our internal evaluation routes, and \tabref{tab:challenge} on our submissions to the CARLA Leaderboard. Note that we could not submit all the baselines from \tabref{tab:baselines} or obtain statistics for multiple evaluations of each model on the Leaderboard due to the limited monthly evaluation budget (200 hours). 

\begin{figure}[t!]
\centering
\includegraphics[width=\columnwidth]{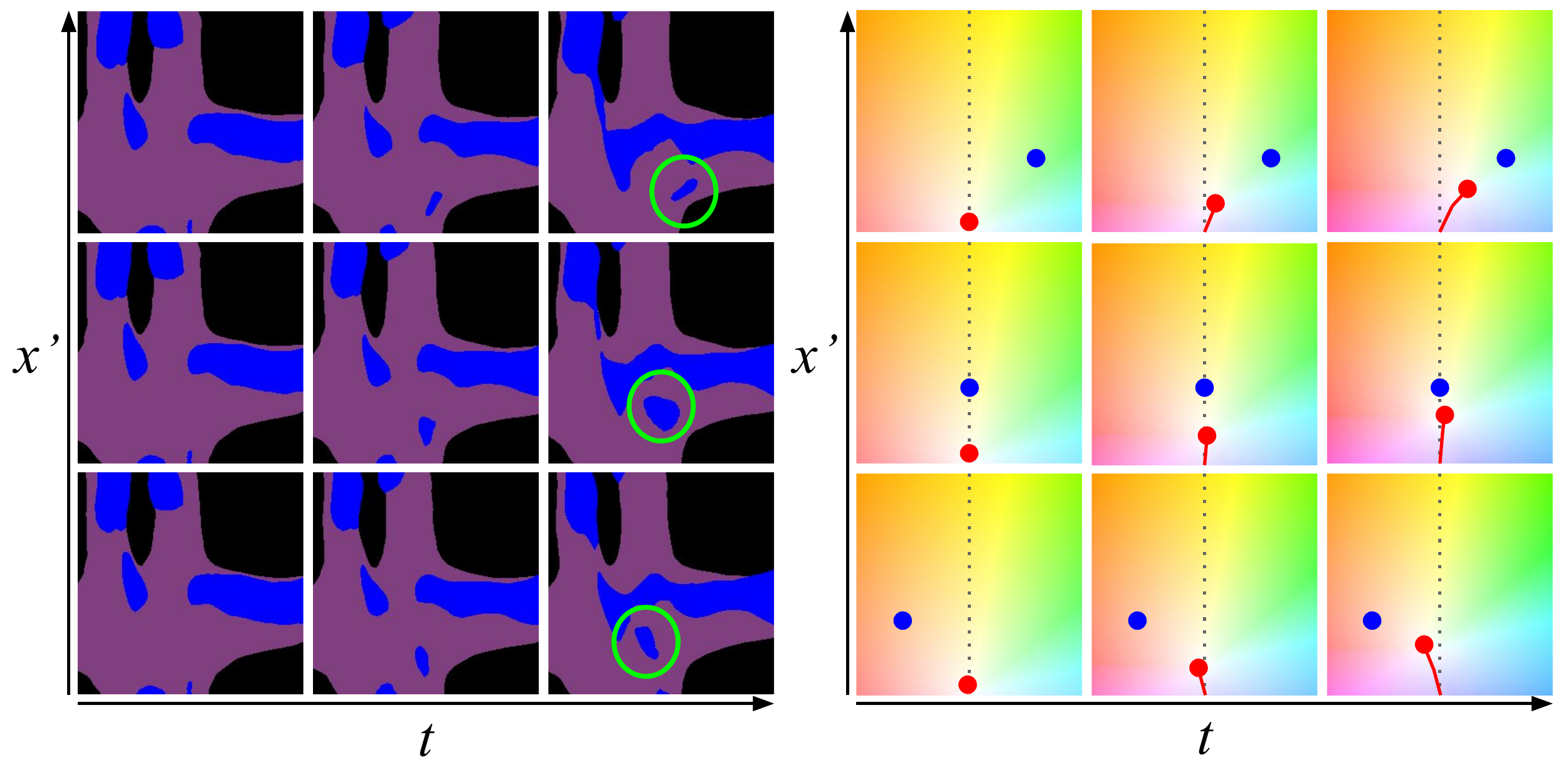}
\vspace{-0.7cm}
\caption{\textbf{NEAT Visualization.} We show $\bs_N$ (left) and $\bo_N$ (right) as we interpolate $x'$ and $t$ for the scene in \figref{fig:teaser}. The {\color{green}green circles} highlight the different predicted ego-vehicle positions. On the right, we show the predicted trajectory and waypoint $\bw_t$ in {\color{red}red}. Note how the model adapts its prediction to the target location $(x',y')$ (shown in {\color{blue}blue}).}
\label{fig:interp}
\vspace{-0.3cm}
\end{figure}

\boldparagraph{Importance of Conditioning}
We observe that in both evaluation settings, CILRS and LBC perform poorly. However, a major improvement can be obtained with a different conditioning signal. CILRS uses discrete navigational commands for conditioning, and LBC uses target locations represented in image space. By using target locations in BEV space and predicting waypoints, AIM and NEAT can more easily adapt their predictions to a change in driver intention, thereby achieving better scores. We show this adaptation of predictions for NEAT in \figref{fig:interp}, by predicting semantics and waypoint offsets for different target locations $x'$ and time steps $t$. The waypoint offset formulation of NEAT introduces a bias that leads to smooth trajectories between consecutive waypoints ({\color{red}red lines} in $\textbf{o}_N$) towards the provided target location in {\color{blue}blue}.

\boldparagraph{AIM-MT and Expert} We observe that AIM-MT is a strong baseline that becomes progressively better with denser forms of auxiliary supervision. The final variant which incorporates dense supervision of both 2D depth and 2D semantics achieves similar performance to NEAT on our 42 internal evaluation routes but does not generalize as well to the unseen routes of the Leaderboard (\tabref{tab:challenge}). Interestingly, in some cases, AIM-MT and NEAT match or even exceed the performance of the privileged expert in \tabref{tab:baselines}. Though our expert is an improved version of the one used in~\cite{Chen2019CORL}, it still incurs some infractions due to its reliance on relatively simple heuristics and driving rules.

\begin{figure}[t]
	\scriptsize
	\setlength{\tabcolsep}{3pt}
	\centering
	\begin{tabular}{c l}
		\multirow{13}{*}{\includegraphics[height=4.15cm]{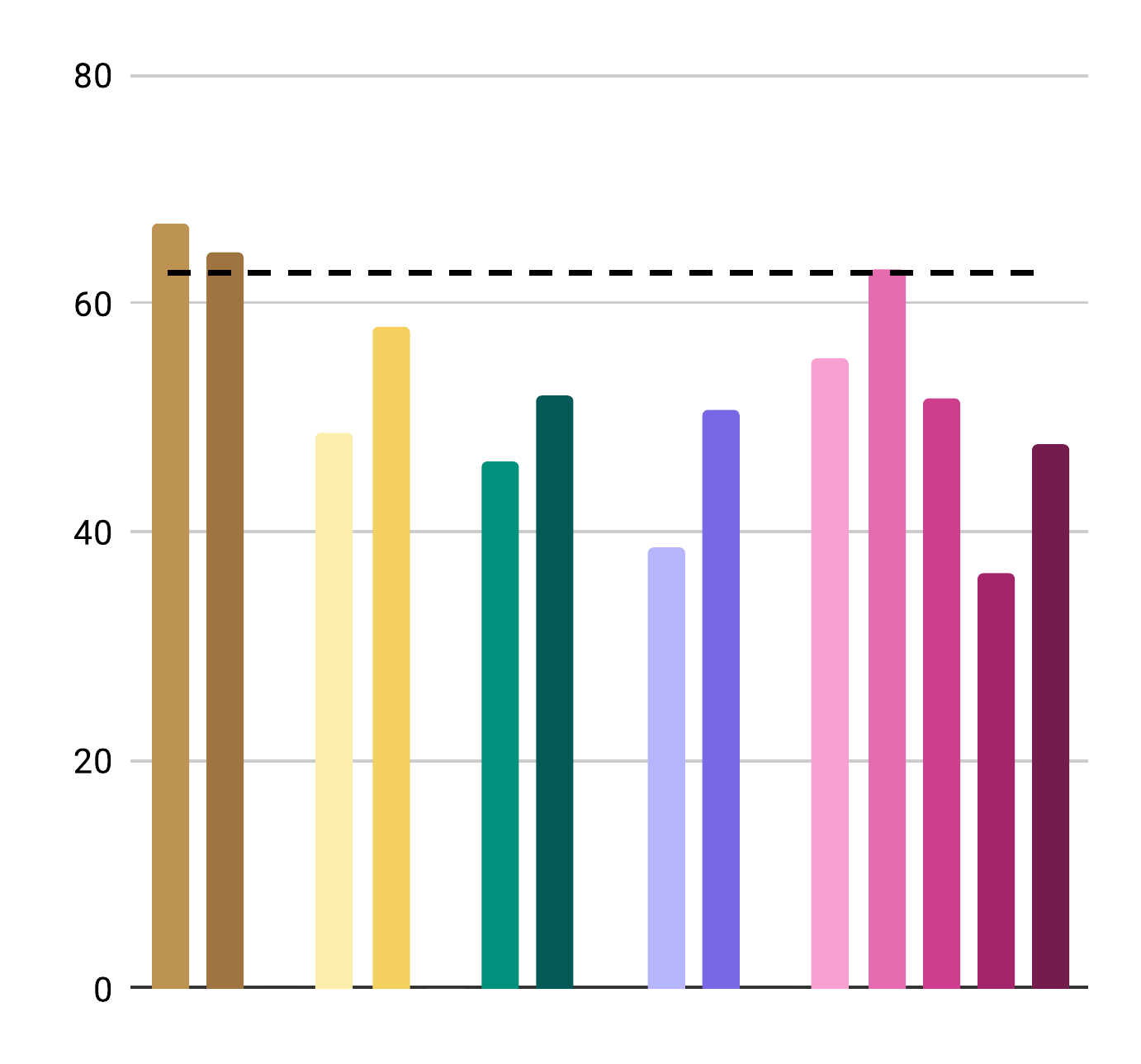}} & \crule[bd9354]{6pt}{6pt} Default Configuration (Seed 1) \\
		& \crule[9e7540]{6pt}{6pt} Default Configuration (Seed 2) \\
		& \crule[fbeeac]{6pt}{6pt} 4 Classes (- Green Light) \\
		& \crule[f4d160]{6pt}{6pt} 6 Classes (+ Lane Marking) \\
		& \crule[00917c]{6pt}{6pt} Less Iterations ($N=1$) \\
		& \crule[025955]{6pt}{6pt} More Iterations ($N=3$) \\
		& \crule[b8b5ff]{6pt}{6pt} Shorter Horizon ($Z=2$) \\
		& \crule[7868e6]{6pt}{6pt} Longer Horizon ($Z=6$) \\
		& \crule[f8a1d1]{6pt}{6pt} No Side Views ($S=1$) \\
		& \crule[e36bae]{6pt}{6pt} No Transformer ($L=0$) \\
		& \crule[cd3e8d]{6pt}{6pt} No Intermediate Loss ($\gamma_1=0$) \\
		& \crule[a4246a]{6pt}{6pt} No Semantic Loss ($\lambda=0$) \\
		& \crule[721d4c]{6pt}{6pt} No Red Light Indicator \\
	\end{tabular}
	\vspace{-0.1cm}
	\caption{\textbf{Ablation Study.} We show the mean DS over our 42 evaluation routes for different NEAT model configurations. Expert performance is shown as a dotted line.}
	\label{fig:ablation}
	\vspace{-0.3cm}
\end{figure}

\begin{figure*}[t!]
\centering
\setlength{\tabcolsep}{1.5pt}
\begin{tabular}{c c c c}
	\multirow{2}{*}[0.08\textwidth]{\includegraphics[height=0.2\textwidth]{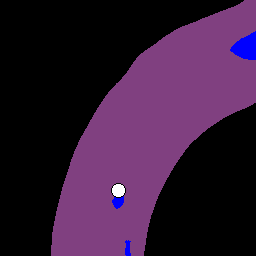}} & \includegraphics[height=0.095\textwidth]{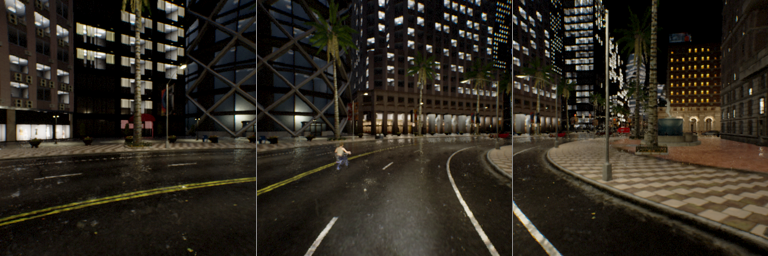} &
	\multirow{2}{*}[0.08\textwidth]{\includegraphics[height=0.2\textwidth]{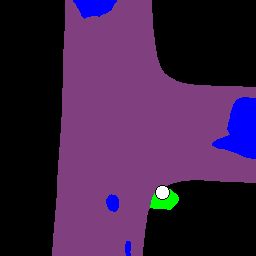}} & \includegraphics[height=0.095\textwidth]{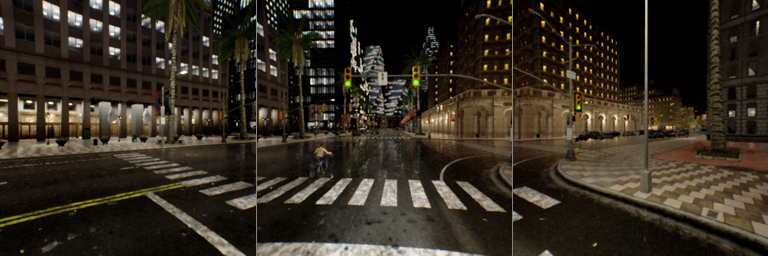} \\
	 & \includegraphics[height=0.095\textwidth]{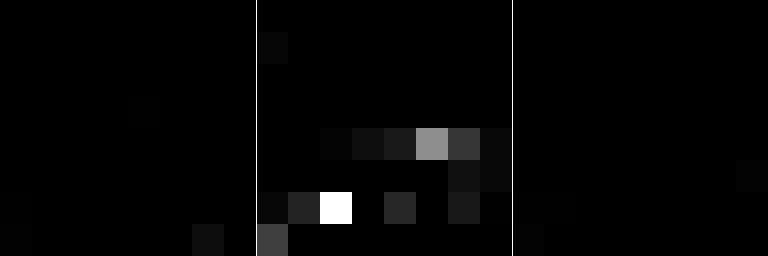} & & \includegraphics[height=0.095\textwidth]{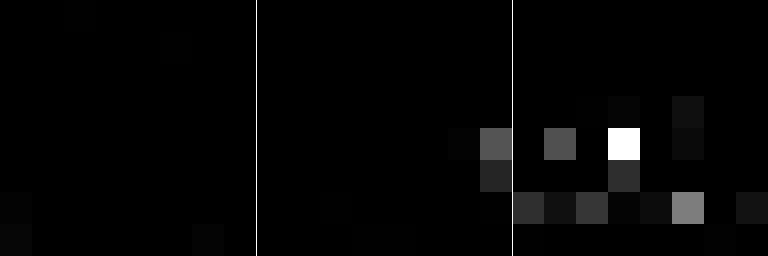} \\
	 \multirow{2}{*}[0.08\textwidth]{\includegraphics[height=0.2\textwidth]{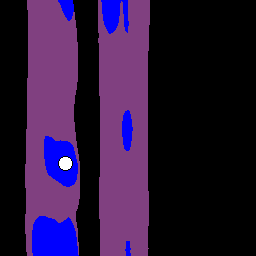}} & \includegraphics[height=0.095\textwidth]{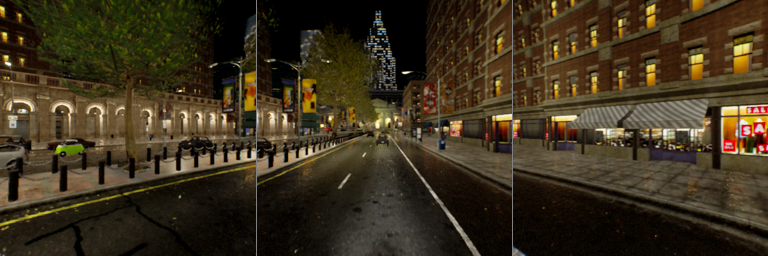} &
	\multirow{2}{*}[0.08\textwidth]{\includegraphics[height=0.2\textwidth]{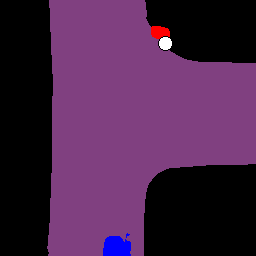}} & \includegraphics[height=0.095\textwidth]{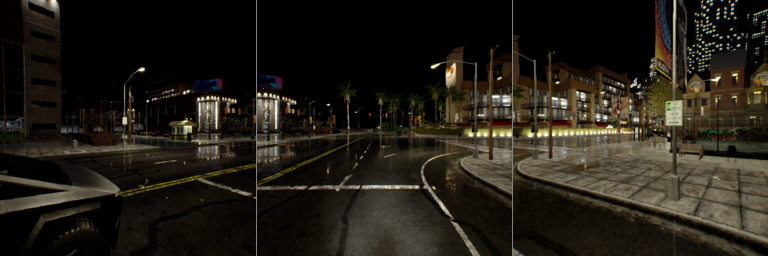} \\
	 & \includegraphics[height=0.095\textwidth]{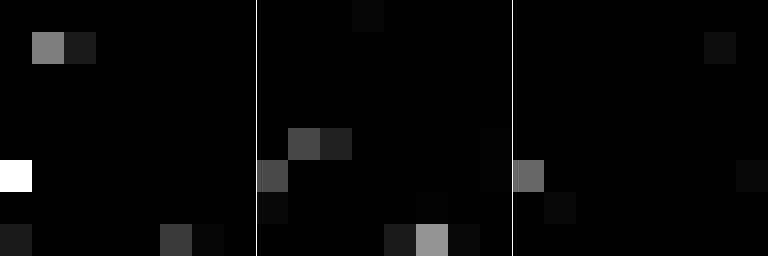} & & \includegraphics[height=0.095\textwidth]{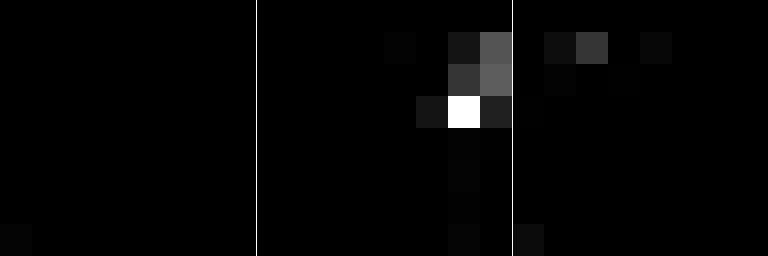} \\
\end{tabular}
\vspace{-0.3cm}
\caption{\textbf{Attention Maps.} We visualize the semantics $\bs_N$ for 4 frames of a driving sequence (legend: {\color{black}none}, {\color{violet}road}, {\color{blue}obstacle}, {\color{red}red light}, {\color{green}green light}). We highlight one particular $(x,y)$ location as a white circle on each $\bs_N$, for which we visualize the input and corresponding attention map $\ba_n$. NEAT consistently attends to the region corresponding to the object of interest (from top left to bottom right: bicyclist, green light, vehicle and red light). Best viewed on screen, zoom in for details.}
\label{fig:att}
\vspace{-0.4cm}
\end{figure*}

\boldparagraph{Leaderboard Results} While NEAT is not the best performing method in terms of DS, it has the safest driving behavior among the top three methods on the Leaderboard, as evidenced by its higher IS. WOR~\cite{Chen2021ARXIV} is concurrent work that supervises the driving task with a Q function obtained using dynamic programming, and MaRLn is an extension of the Reinforcement Learning (RL) method presented in~\cite{Toromanoff2020CVPR}. WOR and MaRLn require 1M and 20M training frames respectively. In comparison, our training dataset only has 130k frames, and can potentially be improved through orthogonal techniques such as DAgger~\cite{Ross2011AISTATS,Prakash2020CVPR}.

\subsection{Ablation Study}

In \figref{fig:ablation}, we compare multiple variants of NEAT, varying the following parameters: training seed, semantic class count ($K$), attention iterations ($N$), prediction horizon ($Z$), input sensor count ($S$), transformer layers ($L$), and loss weights ($\gamma_1,\lambda$). While a detailed analysis regarding each factor is provided in the supplementary, we focus here on four variants in particular: Firstly, we observe that different random training seeds of NEAT achieve similar performance, which is a desirable property not seen in all end-to-end driving models~\cite{Behl2020IROS}. Second, as observed by~\cite{Behl2020IROS}, 2D semantic models (such as AIM-VA and AIM-MT) rely heavily on lane marking annotations for strong performance. We observe that these are not needed by NEAT for which the default configuration with 5 classes outperforms the variant that includes lane markings with 6 classes. Third, in the shorter horizon variant ($Z=2$) with only 2 predicted waypoints, we observe that the output waypoints do not deviate sharply enough from the vertical axis for the PID controller to perform certain maneuvers. It is also likely that the additional supervision provided by having a horizon of $Z=4$ in our default configuration has a positive effect on performance. Fourth, the gain of the default NEAT model compared to its version without the semantic loss ($\lambda=0$) is 30\%, showing the benefit of performing BEV semantic prediction and trajectory planning jointly.

\boldparagraph{Runtime} To analyze the runtime overhead of NEAT's offset prediction task, we now create a hybrid version of AIM and NEAT. This model directly regresses waypoints from NEAT's encoder features $\textbf{c}_0$ using a GRU decoder (like AIM) instead of predicting offsets. We still use a semantic decoder at train time supervised with only the cross-entropy term of Eq. \eqref{eqn:loss}. At test time, the average {runtime} per frame of the hybrid model (with the semantics head discarded) is 15.92 ms on a 3080 GPU. In comparison, the default NEAT model takes 30.37 ms, \ie, both approaches are real-time even with un-optimized code. Without the compute-intensive red light indicator, NEAT's runtime is only 18.60 ms. Importantly, NEAT (DS = 65.10) {significantly outperforms} the AIM-NEAT hybrid model (DS = 33.63). This shows that NEAT's attention maps and location-specific features lead to improved waypoint predictions.

\subsection{Visualizations}
Our supplementary video\footnote{\url{https://www.youtube.com/watch?v=gtO-ghjKkRs}} contains qualitative examples of NEAT's driving capabilities. For the first route in the video, we visualize attention maps $\ba_N$ for different locations on the route in \figref{fig:att}. For each frame in the video, we randomly sample BEV $(x,y)$ locations and pass them through the trained NEAT model until one of the locations corresponds to the class obstacle, red light, or green light. Four such frames are shown in \figref{fig:att}. We observe a common trend in the attention maps: NEAT focuses on the image corresponding to the object of interest, albeit sometimes at a slightly different location in the image. This can be attributed to the fact that NEAT's attention maps are over learned image features that capture information aggregated over larger receptive fields. To quantitatively evaluate this property, we extract the $32\times32$ image patch which NEAT assigns the highest attention weight for one random $(x,y)$ location in each scene of our validation set and analyze its ground truth 2D semantic segmentation labels. The semantic class predicted by NEAT for $(x,y)$ is present in the 2D patch in 79.67\% of the scenes.

\section{Conclusion}
In this work, we take a step towards interpretable, high-performance, end-to-end autonomous driving with our novel NEAT feature representation. Our approach tackles the challenging problem of joint BEV semantic prediction and vehicle trajectory planning from camera inputs and drives with the highest safety among state-of-the-art methods on the CARLA simulator. NEAT is generic and flexible in terms of both input modalities and output task/supervision and we plan to combine it with orthogonal ideas (\eg, DAgger, RL) in the future.

\vspace{0.2cm}
\noindent
\textbf{Acknowledgements:} This work is supported by the BMBF through the T\"ubingen AI Center (FKZ: 01IS18039B) and the BMWi in the project KI Delta Learning (project number: 19A19013O). We thank the International Max Planck Research School for Intelligent Systems (IMPRS-IS) for supporting Kashyap Chitta. The authors also thank Micha Schilling for his help in re-implementing AIM-VA.

{\small
	\bibliographystyle{ieee_fullname}
	\bibliography{bibliography_long,bibliography_custom,bibliography}
}

\end{document}